# Coverage Control for Wire-Traversing Robots*

Gennaro Notomista and Magnus Egerstedt

*Abstract*— In this paper we consider the coverage control problem for a team of wire-traversing robots. The two-dimensional motion of robots moving in a planar environment has to be projected to one-dimensional manifolds representing the wires. Starting from Lloyd's descent algorithm for coverage control, a solution that generates continuous motion of the robots on the wires is proposed. This is realized by means of a Continuous Onto Wires (COW) map: the robots' workspace is mapped onto the wires on which the motion of the robots is constrained to be. A final projection step is introduced to ensure that the configuration of the robots on the wires is a local minimizer of the constrained locational cost. An algorithm for the continuous constrained coverage control problem is proposed and it is tested both in simulation and on a team of mobile robots.

## I. INTRODUCTION

Robots with constrained motion, e.g. those with the ability to move only along pre-designed infrastructures, lend themselves to a large variety of applications, such as environmental monitoring [1] and agricultural robotic tasks [2]. Some of the reasons of their success can be recognized in the following features [1], [3]:

- low energy requirements
- simplicity in the motion control
- small localization errors
- absence of navigation problems even in unknown environments.

However, these advantages are obtained to the detriment of a more complex infrastructure. Nevertheless, there are a lot of applications in which an infrastructure is already present and it can be exploited virtually at no additional cost. An example is power transmission line maintenance [4].

A particular category of constrained motion robots are wire-traversing robots [3], [5]. This paper focuses on the motion planning and control for this kind of robots whose objective is sensor coverage of the surrounding environment.

Wire-traversing robots have already found their application in several domains. In [6] the development of a mobile robot which is able to autonomously navigate on power transmission lines is described. The goal is automating the inspection of power transmission lines and their equipment. Robotics in agriculture and forestry [2] has already experienced an automation process that introduced the use of cable-driven robots whose tasks consist in harvesting fruits and vegetables, dispensing fertilizer and monitoring growth and health of plants. Moving to a different branch, in [7] an algorithm to monitor traffic starting from videos recorded from Skycams ([8]) suggests the viability of wire-traversing autonomous robots for traffic and road network management. In [1], [9] a cable-based robotic platform is described, whose objective is monitoring the environment and characterizing its phenomena. As also pointed out before, the strength of such a system lies in its overall robustness and reliability, accurate and reproducible motion, long range mobility even in complex environments as well as low energy consumption that enables sustainable operation.

Although the technology for the deployment of wire-traversing robots in the environment is somewhat mature, none of the above-mentioned approaches explicitly deals with the motion planning of the robots on the wires on which they are constrained to navigate. In [10] the concept and the design of a mobile manipulator for autonomous installation and removal of aircraft-warning spheres on overhead wires of electric power transmission lines are presented. [11] describes the development of a mobile robot that can navigate aerial power transmission lines autonomously with the goal of automating inspection of power transmission lines. In [12] a multi-unit structure wire mobile robot is proposed, which allows the robot to transfer to a branch wire and avoid obstacles on the wire.

The motion planning for robots on wires is typically left to general purpose motion planners that use search algorithms on grid maps in order to plan a route to a desired location (see e.g. [13]). The main contribution of this paper is a solution to the motion planning problem for wire-traversing robots and, in general, for robots constrained on grid maps. This is achieved by including the motion constraints in the formulation of the motion control law. This concept is applied to a coverage task, where the robots have to spread in the environment in order to monitor its phenomena. Constrained locational optimization has already been considered in [14], where the author proposes a decentralized gradient projection method in order to obtain the motion control law. The advantage on this method will be highlighted in Section II-B. Moreover, a hybrid method, which uses both locational optimization and path planning algorithms to generate robots' motion, is presented in [15].

The remainder of the paper is organized as follows. In Section II, after a brief overview on the Lloyd's algorithm for coverage control for multi-robot systems, the definition of the constrained coverage control problem is introduced and a solution to it is proposed. In Section III the main results of

*This work was sponsored by the U.S. Office of Naval Research through Grant No. N00014-15-2115.

G. Notomista is with the Woodruff School of Mechanical Engineering and the School of Electrical and Computer Engineering, Georgia Institute of Technology, Atlanta, GA 30332 USA, g.notomista@gatech.edu

M. Egerstedt is with the School of Electrical and Computer Engineering, Georgia Institute of Technology, Atlanta, GA 30332 USA, magnus@gatech.edu

the paper are used to synthesize a motion controller for the robots on the wires in order to solve the constrained coverage control problem. In Section IV the derived algorithm is deployed on a team of mobile robots.

## II. COVERAGE ON WIRES

In this section we first introduce the notation that will be used throughout the paper and then derive the results on constrained coverage control. These will be used in Section III to derive the motion control law to be applied to a group of wire-traversing robots.

### A. Locational Optimization

Let $X \subsetneq \mathbb{R}^2$ be a closed and convex polygon and $\partial X$ its boundary. Let $p_1, \ldots, p_N \in \mathbb{R}^2$ denote the locations of $N$ robots moving in the space $X$. We further assume that the motion of the $N$ robots can be modeled using the single integrator dynamics:

$$\dot{p}_i = u_i,$$

where $u_i \in \mathbb{R}^2$ is the control input of robot $i$. Define

$$\mathcal{J}(p_1, \ldots, p_N) = \sum_{i=1}^{N} \int_{V_i} \|x - p_i\|_2^2 \, dx \quad (1)$$

as the locational optimization function [16]. $\mathcal{V}(p_1, \ldots, p_N) = \{V_1, \ldots, V_N\}$ is called a Voronoi partition of the polygon $X$, whose $i$-th Voronoi cell $V_i$ corresponding to robot $i$ is defined as:

$$V_i = \{x \in X \mid \|x - p_i\|_2 \le \|x - p_j\|_2 \ \forall j \ne i\}. \quad (2)$$

The integrand function in the expression of $\mathcal{J}(p_1, \ldots, p_N)$ is an increasing function of the Euclidean norm $\|\cdot\|_2$ and it describes the degradation of the sensing performances of the robots.

The Lloyd's descent algorithm is given by the following motion control law for robot $i$:

$$u_i = k_{\mathrm{p}}(\rho_i - p_i), \quad (3)$$

where $k_{\mathrm{p}} \in \mathbb{R}_+$, and $\rho_i \in \mathbb{R}^2$ is the centroid of the Voronoi cell $V_i$. It is derived by solving the following minimization problem using gradient descent:

$$\min_{p_1, \ldots, p_N} \mathcal{J}(p_1, \ldots, p_N). \quad (4)$$

In [16] the set of critical points of $\mathcal{J}(p_1, \ldots, p_N)$ has been demonstrated to be the set of centroidal Voronoi configurations on $X$ where the location of each robot, $p_i$, is the centroid of the Voronoi cell $V_i$.

### B. Constrained Locational Optimization

In order to describe the constrained motion of the robots on the wires, we define the following function that identifies the $i$-th wire:

$$g_i : x \in X \mapsto a_i^\mathsf{T} x + b_i \in \mathbb{R},$$

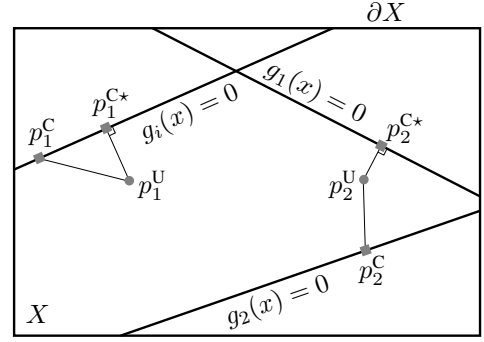

Fig. 1: Robots' workspace $X$ with the $i$-th wire defined by $g_i(x) = 0$. The points $p_1^{\mathrm{U}}$ and $p_2^{\mathrm{U}}$ (gray circles) are the solution of the unconstrained locational optimization problem; the points $p_1^{\mathrm{C}}$ and $p_2^{\mathrm{C}}$ belong to the set $\mathcal{G}$, while $p_1^{\mathrm{C}\star}$ and $p_2^{\mathrm{C}\star}$ (gray squares) are solutions of the minimization problem (7)

where $a_i \in \mathbb{R}^2$ and $b_i \in \mathbb{R}$ for $i = 1, \ldots, N_{\mathrm{w}}$, $N_{\mathrm{w}}$ being the number of wires present in the environment. The wires are then identified as the set:

$$\mathcal{G} = \{x \in X \mid g_i(x) = 0 \text{ for some } i \in \{1, \ldots, N_{\mathrm{w}}\}\} \cup \partial X, \quad (5)$$

where we assume that the boundary of $X$, denoted by $\partial X$, can also be traversed by the robots. The wire-traversing constraint can be formalized as:

$$p_i \in \mathcal{G} \quad \forall i \in \{1, \ldots, N\}. \quad (6)$$

Since the robots are constrained to move on wires, the problem we aim at solving is a constrained version of (4). The integrand function in (1) is non-decreasing therefore, in order to minimize (4), we want to minimize the distance from $p_i$ while remaining on the wires. This can be done by solving the following minimization problem for each robot:

$$\min_{p_i^{\mathrm{C}} \in \mathcal{G}} \|p_i^{\mathrm{C}} - p_i^{\mathrm{U}}\|_2, \quad (7)$$

where $p_1^{\mathrm{U}}, \ldots, p_N^{\mathrm{U}}$ are solutions of (4). The superscripts U and C used in (7) distinguish the positions of the unconstrained robots from those of the wire-constrained robots (see Fig. 1).

The following theorem establishes the equivalence between the minimization problem (4) with the wire-traversing constraints (6) and the minimization problem (7). We say that two minimization problems are equivalent if they have a common local minimizer.

**Theorem 1.** *Given the locational optimization function $\mathcal{J}(p_1, \ldots, p_N)$ defined in (1), the set $\mathcal{G}$ defined in (5) and $p_1^U, \ldots, p_N^U$, solutions of (4), the following minimization problems are equivalent:*

$$\begin{aligned} \min_{p_1, \ldots, p_N} & \ \mathcal{J}(p_1, \ldots, p_N) \\ \text{s.t. } & p_1, \ldots, p_N \in \mathcal{G} \end{aligned} \quad (8)$$

$$\min_{p_i^C \in \mathcal{G}} \|p_i^C - p_i^U\|_2, \quad i = 1, \ldots, N. \quad (9)$$

*Proof.* Let us start proving that (8) ⇒ (9), with which we mean that a solution to (8) is also a solution of (9).
Let $C(p_i) = \prod_{j=1}^{N_w}(a_j^\mathsf{T} p_i + b_j)$. This way we can describe the wire-traversing constraints as follows:

$$p_i \in \mathcal{G} \quad \Leftrightarrow \quad C(p_i) = 0,\ p_i \in X.$$

Writing the Lagrangian for the constrained minimization problem (8), one obtains:

$$L(p_1, \ldots, p_N, \lambda) = J(p_1, \ldots, p_N) + \sum_{i=1}^{N} \lambda_i C(p_i),$$

where $\lambda = [\lambda_1, \ldots, \lambda_N]^\mathsf{T}$ is the Lagrange multiplier. Let $p_1^\star, \ldots, p_N^\star$ be a local minimizer of (8). The following necessary condition has to be satisfied ([17]):

$$\frac{\partial L}{\partial p_i}(p_i^\star) = \frac{\partial J}{\partial p_i}(p_i^\star) + \lambda_i \sum_{k=1}^{N_w} a_k^\mathsf{T} \prod_{\substack{j=1 \\ j \neq k}}^{N_w} (a_j^\mathsf{T} p_i^\star + b_j) = 0 \quad (10)$$

$$\forall i = \{1, \ldots, N\}.$$

Assume that robot $i$ is on wire $\bar{k}$: as a result $a_{\bar{k}}^\mathsf{T} p_i^\star + b_{\bar{k}} = 0$. So, (10) reduces to:

$$\frac{\partial L}{\partial p_i}(p_i^\star) = \frac{\partial J}{\partial p_i}(p_i^\star) + \lambda_i a_{\bar{k}}^\mathsf{T} \prod_{\substack{j=1 \\ j \neq \bar{k}}}^{N_w} (a_j^\mathsf{T} p_i^\star + b_j) = 0$$

$$\forall i = \{1, \ldots, N\}.$$

From [18] we know that $\frac{\partial J}{\partial p_i}(p_i^\star) \parallel (\rho_i - p_i^\star)$, where $\parallel$ is the parallel symbol. Since $\lambda_i \prod_{\substack{j=1 \\ j \neq \bar{k}}}^{N_w}(a_j^\mathsf{T} p_i^\star + b_j) \in \mathbb{R}$ is a scalar, we have that $(\rho_i - p_i^\star) \parallel a_{\bar{k}}$. So, $(\rho_i - p_i^\star)$ is orthogonal to the wire $\bar{k}$ and, therefore, $p_i^\star$ minimizes the distance from $\rho_i$. From [18], we know that $\rho_i$, $i = 1, \ldots, N$ are solutions of (4). Hence, $p_i^\star$ is a local minimizer of (9).

We now prove that (9) ⇒ (8), i.e. a local minimizer of (9) is also a local minimizer of (8). The constraints on (9) are equivalent to $p_i^\mathsf{U} = \rho_i \ \forall i \in \{1, \ldots, N\}$, as shown in [16]. Substituting this expression of $p_i^\mathsf{U}$ in (9), one obtains the following unconstrained minimization problem:

$$\min_{p_i^\mathsf{C} \in \mathcal{G}} \left\| p_i^\mathsf{C} - \rho_i \right\|_2, \quad (11)$$

whose solution $p_i^{\mathsf{C}\star}$ is the closest point to $\rho_i$ that is on the wires defined by $\mathcal{G}$. This means that $(p_i^{\mathsf{C}\star} - \rho_i) \parallel a_k$ for some $k \in \{1, \ldots, N_w\}$. Since also $\frac{\partial J}{\partial p_i}(p_i^{\mathsf{C}\star}) \parallel (p_i^{\mathsf{C}\star} - \rho_i)$, $\exists \lambda_i \in \mathbb{R} \mid \frac{\partial L}{\partial p_i}(p_i^{\mathsf{C}\star}) = 0$. Hence, a solution of (9) is also a local minimizer of (8). □

**Remark 2.** *The equivalence established in Theorem 1 allows us to solve* (9) *instead of* (8)*. What this entails is that, instead of solving a constrained minimization problem, we can solve an unconstrained minimization problem and project its solution onto the constraints' set. Moreover, since we have the motion control law* (3) *that solves the minimization problem* (4)*, we can proceed by just projecting it onto the wires.*

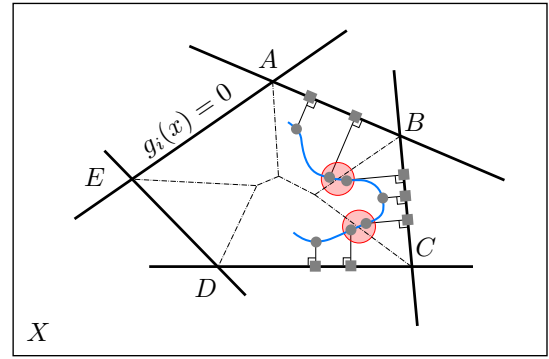

Fig. 2: The areas shaded in red highlight regions where the operator that projects points $p_i^\mathsf{U}$ of the workspace $X$ onto the closest wire is discontinuous. The wires are depicted as thick black lines, while the dash-dot lines represent the medial axis of the polygon $ABCDE$ formed by the wires. The gray circles are points of the blue trajectory that are mapped to the gray squares onto $\mathcal{G}$, the set of wires

Remark 2 points out the advantages of solving an unconstrained optimization problem followed by a projection of the solution onto the constraints set (as in (9)), over solving a constrained minimization problem like (8). However, the set of constraints, $\mathcal{G}$, is the union of affine sets; in fact, each wire is defined as the set $\{x \in X \mid g_i(x) = 0\} = \{x \in X \mid a_i^\mathsf{T} x + b_i = 0\}$, $i \in \{1, \ldots, N_w\}$. So, $\mathcal{G}$ is not convex. As such, the solution of (9) requires optimization methods for non-convex problems. Even though the latter can be also solved efficiently (see [17]), the main objective of the constrained coverage control problem is that of generating a motion control law to be executed by the robots on the wires. A solution to (9), i.e. an *orthogonal projection* onto the wires (as proposed in [14]), does not fulfill this requirement. In Fig. 2 an example of a discontinuous projection on the wires is shown. In particular, the set of discontinuity points coincides with the set of points that have more than one closest point on the set $\mathcal{G}$. It follows that the *medial axes* (or the topological skeletons) of the polygons bounded by the wires, defined as the boundaries of the Voronoi diagrams of the edges of the polygons, are the sets of discontinuity points for the projected motion of the robots onto the wires.

### C. From Projection to Mapping

In this section we describe a method to relax the conditions imposed by Theorem 1 in the interest of producing a continuous motion of the points $p_i$. This will be done by using a Continuous Onto Wires (COW) mapping to the set $\mathcal{G}$ defining the constraints.

Let

$$M : p \in X \subsetneq \mathbb{R}^2 \mapsto p^M \in \mathcal{G} \subsetneq \mathbb{R}^2 \quad (12)$$

be the operator that maps the robot workspace $X$ to the set of

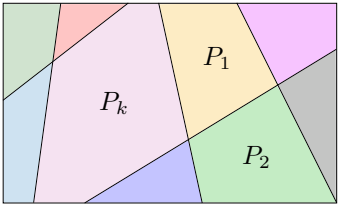

Fig. 3: The polygonal tessellation induced by the wires consists of all closed and convex polygons $P_k$ as they results from the intersection of half planes

wires $\mathcal{G}$, where $p^M$ does not necessarily solve the program:

$$\min_{x \in \mathcal{G}} \|x - p\|_2.$$

In order to find a suitable expression for such a mapping, elements from complex analysis will be required. For this reason they are recalled in the following.

First of all, let us define the following isomorphism between $\mathbb{R}^2$ and $\mathbb{C}$:

$$\mathcal{I} : \begin{bmatrix} a \\ b \end{bmatrix} \in \mathbb{R}^2 \mapsto a + \iota b \in \mathbb{C}, \quad (13)$$

where $\iota$ is the imaginary unit. Moreover, in the following we will use $\Re(\cdot)$ to denote the operator that extracts the real part of a complex number. With abuse of notation, we denote with $\widetilde{X} = \mathcal{I}(X) \subsetneq \mathbb{C}$ the image of the robots' workspace $X$ through the isomorphism (13). Let $\widetilde{p}_1, \ldots, \widetilde{p}_N \in \widetilde{X} \subsetneq \mathbb{C}$ be the robots' positions in the complex plane. Similarly, we can define $\widetilde{g}_i$ and $\widetilde{\mathcal{G}}$.

The following results will be used in the definition of a mapping (12).

**Definition 3** (Conformal map [19]). *Let $X_1$ and $X_2$ be two open subsets of $\mathbb{C}$. A map $f : X_1 \to X_2$ is said to preserve angles if for every two differentiable curves $\gamma_1 : t \in [-\epsilon, \epsilon] \subsetneq \mathbb{R} \mapsto c \in \mathbb{C}$ and $\gamma_2 : t \in [-\epsilon, \epsilon] \subsetneq \mathbb{R} \mapsto c \in \mathbb{C}$, where $\gamma_1(0) = \gamma_2(0) = c^\star$, the angle formed by their tangents at $c^\star$ is the same as the angle formed by the tangents of the mapped curves $f \circ \gamma_1$ and $f \circ \gamma_2$ at $f(c^\star)$. A conformal map from $X_1$ to $X_2$ is a differentiable bijection that preserves angles.*

With this definition, we can now state the following theorem.

**Theorem 4** (Riemann mapping theorem [20]). *Let $X \subsetneq \mathbb{C}$ be a simply connected region of the complex plane, and let $x \in X$. Then, there is a unique conformal map $f : X \to \mathbb{D}$, where $\mathbb{D}$ is the unit disc, such that $f(x) = 0$ and $f'(x) = 0$.*

From this theorem the following corollary can be proven.

**Corollary 5.** *Two simply connected regions of the complex plane, $X_1, X_2 \subsetneq \mathbb{C}$, are homeomorphic.*

**Fact 6.** *The wires defined by the set $\widetilde{\mathcal{G}}$ induce a polygonal tessellation of $\widetilde{X}$. The resulting polygonal areas $P_k$ are closed and convex as they come from the intersection of half planes (see Fig. 3).*

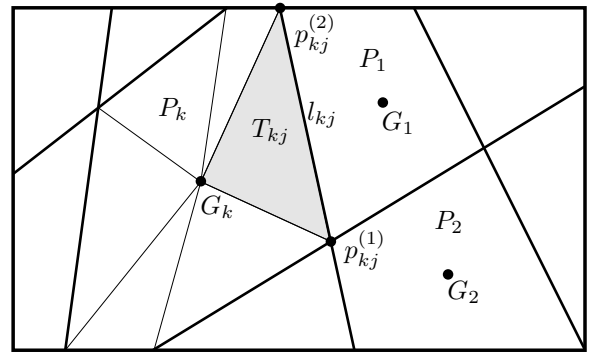

Fig. 4: Quantities used in the formulation of a COW map $\widetilde{M}$

Using the result of Corollary 5, we can construct a mapping from each of the polygons of the polygonal tessellation introduced in Fact 6 onto their boundaries, i.e. the wires. This can be realized as follows.

Let $P_k \subseteq \widetilde{X} \subsetneq \mathbb{C}$, $k = 1, \ldots, K$ be the $K$ polygons resulting from the polygonal tessellation defined by the wires $\widetilde{\mathcal{G}}$, and let $G_k \in \mathbb{C}$, $k = 1, \ldots, K$ be their corresponding centroids. Let $l_{kj} \subsetneq \widetilde{X} \subsetneq \mathbb{C}$, $j = 1, \ldots, J$ denote the $J$ sides of the polygon $P_k$, and $p_{kj}^{(1)} \in \mathbb{C}$ and $p_{kj}^{(2)} \in \mathbb{C}$ the two endpoints of the side $l_{kj}$. Note that $\widetilde{\mathcal{G}}$, the subset of the complex plane isomorphic to $\mathcal{G}$ through (13), indicated with abuse of notation by $\mathcal{I}(\mathcal{G})$, can be defined as:

$$\widetilde{\mathcal{G}} = \mathcal{I}(\mathcal{G}) = \bigcup_{k=1}^{K} \bigcup_{j=1}^{J} l_{kj}. \quad (14)$$

Let us define $T_{kj} \subsetneq \widetilde{X} \subsetneq \mathbb{C}$ as the triangle with vertices $p_{kj}^{(1)}$, $p_{kj}^{(2)}$ and $G_k$. Fig. 4 shows all the quantities that have been just introduced.

With these premises, let us define the mappings $m_{kj}$ as follows:

$$m_{kj} = \begin{cases} T_{kj} \subsetneq \widetilde{X} \subsetneq \mathbb{C} \to l_{kj} \subsetneq \widetilde{X} \subsetneq \mathbb{C} \\ \widetilde{X} \setminus T_{kj} \subsetneq \widetilde{X} \subsetneq \mathbb{C} \to \{0\} \subsetneq \mathbb{C} \end{cases}.$$

So, the mapping

$$\widetilde{M} : x \in \widetilde{X} \subsetneq \mathbb{C} \mapsto \sum_{k=1}^{K} \sum_{j=1}^{J} m_{kj}(x) \in \widetilde{\mathcal{G}} \subsetneq \mathbb{C} \quad (15)$$

transforms the robot workspace $\widetilde{X} \subsetneq \mathbb{C}$ to the set of wires $\widetilde{\mathcal{G}} \subsetneq \mathbb{C}$ in the complex plane.

**Remark 7.** *Let $x \in \mathbb{C}$ and $p \in \mathbb{R}^2$ such that $\mathcal{I}(p) = x$. By the properties of the isomorphism (13) and by the definitions of the mappings (12) and (15), we characterize $M$ by the following equations:*

$$\widetilde{M}(x) = \mathcal{I}\left(M\left(\mathcal{I}^{-1}(x)\right)\right)$$
$$M(p) = \mathcal{I}^{-1}\left(\widetilde{M}\left(\mathcal{I}(p)\right)\right).$$

Now the expression of $m_{kj}$ is left to define. In order to do so, let us introduce a particular conformal mapping.

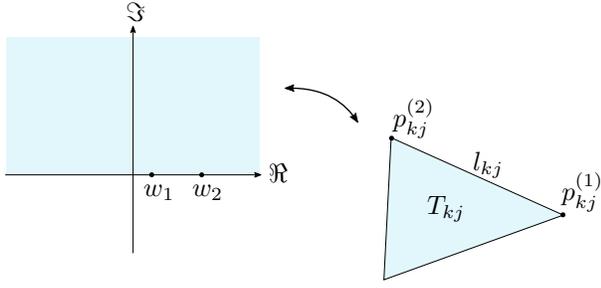

Fig. 5: Schwar-Christoffel mapping between the upper-half plane $\mathbb{H}$ and the triangular region $T_{kj}$ of the complex plane. The prevertices $w_1$ and $w_2$ are mapped to the vertices of the triangle $p_{kj}^{(1)}$ and $p_{kj}^{(2)}$, respectively

**Definition 8** (Schwarz-Christoffel mapping [21]). *A Schwarz-Christoffel mapping is a conformal mapping from the upper half-plane $\mathbb{H} = \{x \in \mathbb{C} \mid \Re(x) \geq 0\}$ (the canonical domain) to a region $\mathbb{P}$ of the complex plane bounded by a polygon (the physical domain). Its expression is given by:*

$$f : x \in \mathbb{H} \mapsto x_0 + c \int_{x_0}^{x} \prod_{j=1}^{J-1} (\chi - w_j)^{\alpha_j - 1} \, d\chi \in \mathbb{P},$$

*where $x_0, c \in \mathbb{C}$ are two constants that translate, rotate and scale the polygon that bounds $\mathbb{P}$, $J$ is the number of sides of the polygon, $\alpha_j$ is the interior angle at the $j$-th vertex of the polygon and $w_j \in \mathbb{R}$, $j = 1, \ldots, J-1$ are called the prevertices and have the property of being mapped to the vertices of the polygon. In case of triangular domains, i.e. $J = 3$, the prevertices can be arbitrarily set to any location on the real axis.*

**Definition 9.** *Let*

$$f_{kj} : \mathbb{H} \to T_{kj} \quad (16)$$

*be the Schwarz-Christoffel mapping between the upper-half plane and the triangular region $T_{kj}$ defined above. The real axis is mapped to the boundary of $T_{kj}$ and the prevertices $w_1$ and $w_2$ are mapped to $p_{kj}^{(1)}$ and $p_{kj}^{(2)}$, respectively (see Fig. 5). Because of the fact that for triangular physical domains the prevertices can be arbitrarily chosen on the real axis, we assume that their value does not depend on the indexes $k$ and $j$.*

The following theorem defines a COW map, i.e. a continuous and onto mapping, from the robots' workspace $\widetilde{X}$ to the set $\widetilde{\mathcal{G}}$ defined in (14).

**Theorem 10.** *Let $f_{kj}$ be the Schwarz-Christoffel mapping defined by (16), $w_1$ and $w_2$ the prevertices of the mapping $f_{kj}$; let $p_{kj}^{(1)}$, $p_{kj}^{(2)}$ and $G_k$ the vertices of the triangular boundary of the region $T_{kj}$. The mapping $\widetilde{M}$ defined in (15)*

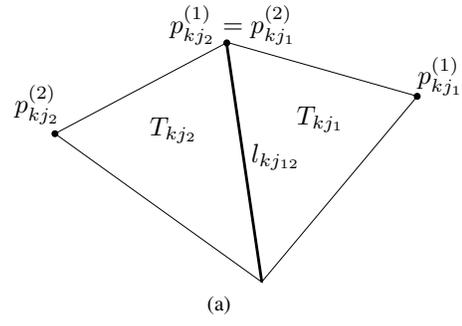

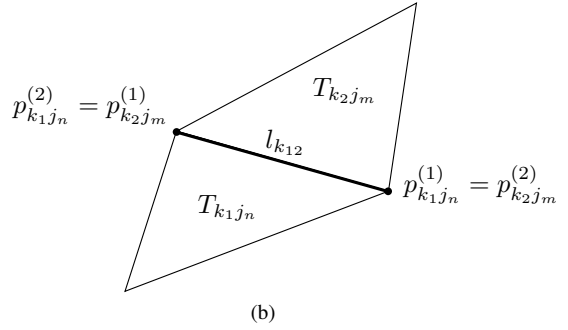

(a)

(b)

Fig. 6: Adjacent triangular regions over whose common edges the continuity of the function $\widetilde{M}$ has to be shown

*where $m_{kj}$ is given by the following onto mapping:*

$$m_{kj}(x) = \begin{cases} p_{kj}^{(1)} & \text{if } \Re\left(f_{kj}^{-1}(x)\right) \leq w_1 \\ f_{kj}\left(\Re\left(f_{kj}^{-1}(x)\right)\right) & \text{if } w_1 < \Re\left(f_{kj}^{-1}(x)\right) < w_2 \\ p_{kj}^{(2)} & \text{if } \Re\left(f_{kj}^{-1}(x)\right) \geq w_2 \end{cases},$$

*defines a continuous mapping from the robots' workspace $\widetilde{X} \setminus \left(\bigcup_{k=1}^{K} \{G_k\}\right) \subsetneq \mathbb{C}$ to the set $\widetilde{\mathcal{G}} \subsetneq \mathbb{C}$ defined in (14).*

*Proof.* By the Definition 3 of conformal map, $f^{-1}$ exists and it is continuous as it is the map $f$ itself. As the operator $\Re(\cdot)$ is continuous and the composition of continuous functions is continuous, one has that $m_{kj}$ is continuous for the vertical strip of the complex plane defined by $w_1 < \Re\left(f^{-1}(x)\right) < w_2$. For $\Re\left(f^{-1}(x)\right) \leq w_1$ and $\Re\left(f^{-1}(x)\right) \geq w_2$, $m_{kj}$ is constant and so continuous. For $\Re\left(f^{-1}(x)\right) = w_1$, $f\left(\Re\left(f^{-1}(x)\right)\right) = f(w_1) = p_{kj}^{(1)}$ by Definition 9 of $f_{kj}$. Hence $m_{kj}$ is continuous on the vertical line of the complex plane defined by $\Re\left(f^{-1}(x)\right) = w_1$. A similar argument holds for when $\Re\left(f^{-1}(x)\right) = w_2$. Hence, the mapping $\widetilde{M}$ is continuous over each triangular domain.

Now the continuity of $m_{kj}$ across adjacent domains $T_{kj_1}$ and $T_{kj_2}$ or $T_{k_1 j_n}$ and $T_{k_2 j_m}$ is left (see Fig. 6). Let us define $l_{kj_{12}}$ to be the common segment of the two adjacent regions $T_{kj_1}, T_{kj_2} \subsetneq P_k$ (see Fig. 6a). For $x \in l_{kj_{12}}$ one has that $\Re\left(f_{kj_1}^{-1}(x)\right) \geq w_2$ and $\Re\left(f_{kj_2}^{-1}(x)\right) \leq w_1$. Therefore, in the former case $x$ is mapped to $p_{kj_1}^{(2)}$, whilst in the latter case $x$ is mapped to $p_{kj_2}^{(1)}$. The two points coincide, hence $\widetilde{M}$ is continuous on $l_{kj_{12}}$. Let us now define $l_{k_{12}}$ the common segment of the two adjacent regions $T_{k_1 j_n} \subsetneq P_{k_1}$ and

$T_{k_2 j_m} \subsetneq P_{k_2}$, that is also the only common segment between the two polygons $P_{k_1}$ and $P_{k_2}$ (see Fig. 6b). By Definition 9, for $x \in l_{k_{12}}$ one has $\Re\left(f_{k_1 j_n}^{-1}(x)\right) = f_{k_1 j_n}^{-1}(x)$ and $\Re\left(f_{k_2 j_m}^{-1}(x)\right) = f_{k_2 j_m}^{-1}(x)$. Since $w_1 < \Re\left(f_{k_1 j_n}^{-1}(x)\right) < w_2$ and $w_1 < \Re\left(f_{k_2 j_m}^{-1}(x)\right) < w_2$, we can write:

$$m_{k_1 j_n}(x) = f_{k_1 j_n}\left(\Re\left(f_{k_1 j_n}^{-1}(x)\right)\right) = f_{k_1 j_n}\left(f_{k_1 j_n}^{-1}(x)\right) = x$$

$$m_{k_2 j_m}(x) = f_{k_2 j_m}\left(\Re\left(f_{k_2 j_m}^{-1}(x)\right)\right) = f_{k_2 j_m}\left(f_{k_2 j_m}^{-1}(x)\right) = x.$$

So $m_{k_1 j_n}(x) = m_{k_2 j_m}(x) \; \forall x \in l_{k_{12}}$. Hence, $\widetilde{M}$ is continuous on $l_{k_{12}}$. □

**Remark 11.** *Solving an optimization problem such as* (11) *performs a projection operation in the physical domain* $\mathbb{P}$ *of a Schwarz-Christoffel mapping. The same result is obtained in the canonical domain* $\mathbb{H}$ *by means of the operator* $\Re(\cdot)$.

### III. MOTION CONTROL ON THE WIRES

The COW map $\widetilde{M}$ defined by Theorem 10 allows the direct derivation of a motion control law to be executed by each of the robots on the wires. The resulting motion is continuous and inherently takes into account the constraints defined by the wires.

#### A. Mapped Gradient Descent

As the motion control law derived in this section are to be applied to all the robots without distinction, the subscript $i$ will be dropped from now on.

Note also that all the quantities used in the following are complex numbers, subsets of the complex plane and complex-valued functions of complex variables. Due to the isomorphism (13), the formulation in $\mathbb{C}$ and that in $\mathbb{R}^2$, even though formally different, are substantially equivalent.

The COW mapping derived in Theorem 10 that transforms the domain $\widetilde{X}$ into $\widetilde{G}$ is denoted by:

$$x^{\widetilde{M}} = \widetilde{M}(x), \tag{17}$$

where $\widetilde{M}$ is defined in (15). Differentiating (17), one obtains:

$$\dot{x}^{\widetilde{M}} = \frac{\partial \widetilde{M}}{\partial x} \dot{x} = \frac{\partial \widetilde{M}}{\partial x} k_p (\widetilde{\rho} - x), \tag{18}$$

where $\widetilde{\rho} = \mathcal{I}(\rho)$ and $\dot{x} = k_p(\widetilde{\rho} - x)$ comes from (3).

As far as the expression of $\frac{\partial \widetilde{M}}{\partial x}$ is concerned, starting from (15), it can be written as:

$$\frac{\partial \widetilde{M}}{\partial x} = \sum_{k=1}^{K} \sum_{j=1}^{J} \frac{\partial m_{kj}}{\partial x}.$$

It has to be noticed that, since the operator $\Re(\cdot)$ is not differentiable, $\frac{\partial m_{kj}}{\partial x}$ is not well-defined. However, since we are interested in deriving a motion control law for the robots on the wires, we actually need only the directional derivative of $\widetilde{M}$ and $m_{kj}$ along the wires, denoted by $\partial_{\widetilde{G}} \widetilde{M}$ and $\partial_{\widetilde{G}} m_{kj}$, respectively. Consequently, we need the directional derivative of $\Re$ only along the real axis. The latter is well-defined and it is identically equal to 1. Thus, we can write:

$$\partial_{\widetilde{G}} m_{kj} = \begin{cases} 0 & \text{if } \Re\left(f_{kj}^{-1}(x)\right) \leq w_1 \\ m'_{kj} & \text{if } w_1 < \Re\left(f_{kj}^{-1}(x)\right) < w_2 \\ 0 & \text{if } \Re\left(f_{kj}^{-1}(x)\right) \geq w_2 \end{cases},$$

where $m'_{kj}$ is given by

$$m'_{kj} = \left(\frac{\partial f_{kj}}{\partial x} \circ \Re\left(f_{kj}^{-1}(x)\right)\right) \left(\frac{\partial \Re}{\partial x} \circ f_{kj}^{-1}(x)\right) \frac{\partial f_{kj}^{-1}}{\partial x} =$$

$$= \left(\frac{\partial f_{kj}}{\partial x} \circ \Re\left(f_{kj}^{-1}(x)\right)\right) \left(1 \circ f_{kj}^{-1}(x)\right) \frac{\partial f_{kj}^{-1}}{\partial x}$$

$$= \left(\frac{\partial f_{kj}}{\partial x} \circ \Re\left(f_{kj}^{-1}(x)\right)\right) \frac{1}{\frac{\partial f_{kj}}{\partial x}},$$

where

$$\frac{\partial f_{kj}}{\partial x} = \prod_{i=1}^{2} (x - w_i)^{\alpha_i - 1},$$

and all the quantities used here and specified in Definition 8 are specific for the triangular region $T_{kj}$.

The theorem stated below follows directly from the derivation of (18).

**Theorem 12.** *Let $x_i = \mathcal{I}(p_i)$, $i \in \{1, \ldots, N\}$ be the positions of $N$ robots expressed as points of the complex plane $\mathbb{C}$. Let*

$$\widetilde{\mathcal{J}}(x_1, \ldots, x_N) = \mathcal{I}\left(\mathcal{J}\left(\mathcal{I}^{-1}(x_1), \ldots, \mathcal{I}^{-1}(x_N)\right)\right) \tag{19}$$

*be the locational cost defined based on* (1). *The motion control law*

$$\dot{x}^{\widetilde{M}} = \partial_{\widetilde{G}} \widetilde{M} \, \dot{x}_{\widetilde{G}}, \tag{20}$$

*where the subscript $\widetilde{G}$ indicates quantities mapped onto the wires, applied to robots whose motion is constrained to be in the set $\widetilde{G} = \mathcal{I}(G) \subsetneq \mathbb{C}$ defined in* (14)*, solves the constrained optimization problem:*

$$\begin{aligned} \min_{x_1, \ldots, x_N} & \widetilde{\mathcal{J}}(x_1, \ldots, x_N) \\ \text{s.t.} & \widetilde{M}(x_i) = x_i \quad \forall i \in \{1, \ldots, N\} \end{aligned}. \tag{21}$$

#### B. From Mapping to Projection

In Section II-C, in order to derive a motion control law that ensures a continuous motion of the robots on the wires, we relaxed the constraints imposed in the optimization problems defined in Theorem 1. Now we propose an algorithm that restores those constraints and ensures the best coverage quality that is achievable when the robots are constrained to move on wires.

Let us start stating the following result.

**Fact 13.** *Employing the Euclidean distance as a metric to measure the distance $d(x, l)$ between a point $x$ and a segment $l$, required for the evaluation of Voronoi partitions, the Voronoi cells determined by the edges of a convex polygon*

*are convex polygons themselves. This also means that the medial axis of a convex polygon consists of all straight lines.*

This allows the definition of conformal mappings, similar to $f_{kj}$ introduced in (16), between the upper-half plane $\mathbb{H}$ and the convex and polygonal Voronoi cells related to polygon $P_k$.

With the objective of mapping the point $x \in X$ to its closest wire, each triangular region $T_{kj}$ of polygon $P_k$ is continuously deformed to its corresponding Voronoi cell

$$V_{kj} = \{x \in P_k \mid \mathrm{d}(x, l_{kj}) \leq \mathrm{d}(x, l_{k\bar{j}}) \quad \forall \bar{j} \neq j\}, \quad (22)$$

with which it shares the side $l_{kj}$. See Fig. 7a to Fig. 7e.

The optimization problem (21) is solved using gradient descent that results in the control law (18) applied to each robot. Let us define $\tau_f$ as the time instant at which the velocities given by $k_\mathrm{p}(\widetilde{\rho} - x)$ become sufficiently small for all the robots. For a given time interval $[\tau_f, \tau_f + \tau]$, let

$$\mathcal{D}_{kj} : t \in [\tau_f, \tau_f + \tau] \mapsto T_{kj}^{(t)}, \quad (23)$$

with

$$\mathcal{D}_{kj}(\tau_f) = T_{kj}^{(\tau_f)} = T_{kj}$$
$$\mathcal{D}_{kj}(\tau_f + \tau) = T_{kj}^{(\tau_f + \tau)} = V_{kj},$$

be the deformation operator that transforms the region $T_{kj}$ into the Voronoi cell $V_{kj}$ during the time interval $[\tau_f, \tau_f + \tau]$.

Following the definition in (15), let us define the following COW map at time $t \in [\tau_f, \tau_f + \tau]$:

$$\widetilde{M}^{(t)} : x \in \widetilde{X} \subsetneq \mathbb{C} \mapsto \sum_{k=1}^{K} \sum_{j=1}^{J} m_{kj}^{(t)}(x) \in \widetilde{\mathcal{G}} \subsetneq \mathbb{C},$$

where

$$m_{kj}^{(t)} = \begin{cases} T_{kj}^{(t)} \subsetneq \widetilde{X} \subsetneq \mathbb{C} \to l_{kj} \subsetneq \widetilde{X} \subsetneq \mathbb{C} \\ \widetilde{X} \setminus T_{kj}^{(t)} \subsetneq \widetilde{X} \subsetneq \mathbb{C} \to \{0\} \subsetneq \mathbb{C} \end{cases}.$$

The velocity $\dot{x}^{\widetilde{M}^{(t)}}$ is evaluated using (20) where $\widetilde{M}^{(t)}$ is used in place of $\widetilde{M}$.

**Remark 14.** *By definition of Voronoi cell (22), once the transformation (23) is completed, i.e. $t = \tau_f + \tau$, the COW map $\widetilde{M}^{(\tau_f + \tau)}$ transforms each point $x$ to a point belonging to $\widetilde{\mathcal{G}}$ on the closest wire. Therefore, the robots can execute gradient descent on the wire on which they are at time $t \geq \tau_f + \tau$ in order to get to the positions that minimize (19).*

Algorithm 1 outlines the motion control strategy executed by each robot on the wires. The resulting behavior is depicted in Fig. 8a to Fig. 8d.

Based on the derivation of Algorithm 1, we can state the following theorem.

**Theorem 15.** *Algorithm 1 solves the following constrained optimization problem:*

$$\begin{aligned} \min_{x_1, \ldots, x_N} & \widetilde{\mathcal{J}}(x_1, \ldots, x_N) \\ \text{s.t. } & x_1, \ldots, x_N \in \widetilde{\mathcal{G}} \end{aligned}. \quad (24)$$

---

**Algorithm 1** Continuous Constrained Coverage Control

**Require:** $x$ (robot initial position), $\widetilde{M}$, $\widetilde{M}^{(t)}$, $\mathcal{D}_{kj}$
**Ensure:** Continuous Constrained Coverage Control
  **while** $|k_\mathrm{p}(\widetilde{\rho} - x)| \geq \epsilon$ **do**
    $\widetilde{u} \leftarrow$ compute $\dot{x}^{\widetilde{M}}$
    execute $\widetilde{u}$
  **end while**
  **for** $t = \tau_f$ **to** $\tau_f + \tau$ **do**
    **for** all adjacent regions $T_{kj}$ **do**
      $T_{kj}^{(t)} \leftarrow \mathcal{D}_{kj}(t)$
      $m_{kj}^{(t)} \leftarrow$ compute $m_{kj}^{(t)}$
    **end for**
    $\widetilde{M}^{(t)} \leftarrow \sum_{k=1}^{K} \sum_{j=1}^{J} m_{kj}^{(t)}(x)$
    $\widetilde{u} \leftarrow$ compute $\dot{x}^{\widetilde{M}^{(t)}}$
  **end for**
  execute gradient descent on the current wire

---

## IV. EXPERIMENTS

The algorithm to execute the continuous constrained coverage control described in Algorithm 1 has been deployed on a swarm of ground mobile robots on the Robotarium, a remotely accessible swarm robotics testbed [22], where the robots have been artificially constrained to move on wires.

The algorithm has been implemented in MATLAB and submitted through the Robotarium web interface[1] in order to be executed on the real robots.

Fig. 9a to Fig. 9e show images taken from the video recorded during the experiments. An overhead projector visualizes the wires on which the robots are constrained (thick gray lines). The virtual robots projected on the testbed are linked to the real ones by means of the mappings $\widetilde{M}$ and $\widetilde{M}^{(t)}$ and they move in order to minimize the unconstrained locational cost (1). In Fig. 9a the robots are initialized to random positions on the wires. In Fig. 9b and Fig. 9c the robots execute the control law (20) moving on the wires until the velocities of the projected robots are below a minimum threshold (Fig. 9d). At this point the deformation (23) is performed, and executing gradient descent on the wires on which the robots are at time $\tau_f + \tau$ brings them to the final positions that solve the constrained locational optimization problem (24).

## V. CONCLUSIONS

In this paper we propose a solution to the coverage control problem for wire-traversing robots. Starting from coverage control derived for robots moving in a planar environment, the resultant two-dimensional motion is mapped, in a continuous fashion, onto one-dimensional manifolds which represent the wires. The main contribution of this paper is the derivation of a continuous motion control law that is to be executed by the robots on the wires in order to minimize the constrained locational cost. This is realized by defining a Continuous Onto Wires (COW) map that continuously maps

---
[1] http://www.robotarium.org/

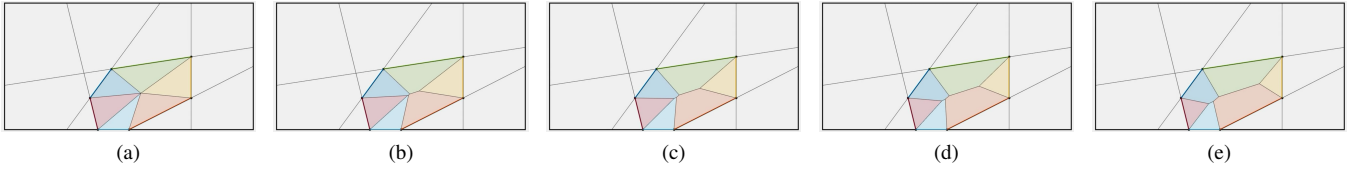

Fig. 7: Continuous deformation of the triangular regions $T_{kj}$ (a) to the corresponding Voronoi cells $V_{kj}$ (e)

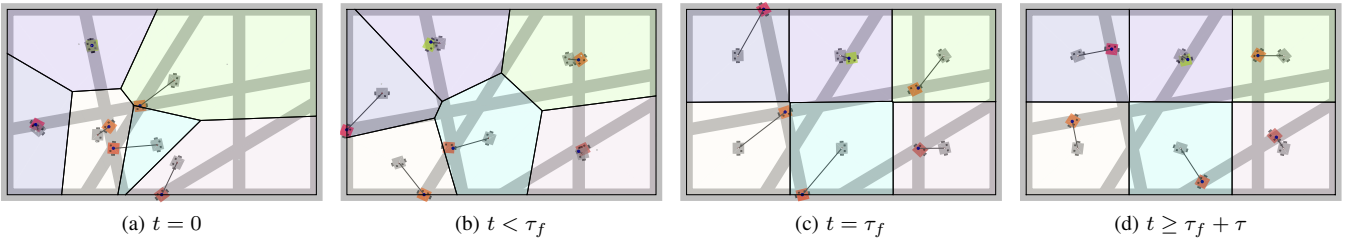

(a) $t = 0$  (b) $t < \tau_f$  (c) $t = \tau_f$  (d) $t \geq \tau_f + \tau$

Fig. 8: Motion of the robots under coverage control constrained by the wires resulting by the application of Algorithm 1. The thick lines are the wires that constrain the motion of the colored robot. The gray robots move according to the control law derived from the minimization of the locational cost (1). The colored area represent the Voronoi cells (2) related to the gray robots. Each gray robot is linked to the corresponding colored robot on the wires to which it is mapped

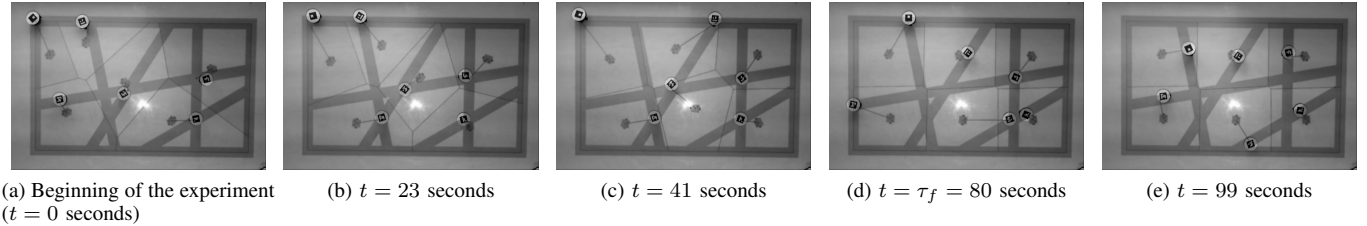

(a) Beginning of the experiment ($t = 0$ seconds)  (b) $t = 23$ seconds  (c) $t = 41$ seconds  (d) $t = \tau_f = 80$ seconds  (e) $t = 99$ seconds

Fig. 9: Algorithm 1 is deployed on a team of robots on the Robotarium. An overhead projector is visualizing information related to the experiment: the thick lines are the wires on which the real robots are constrained to move, the motion of the the projected robots is determined by solving the minimization problem (4), the thin lines are the boundary of the Voronoi cells (2). As in Fig. 8, the projected robots are linked with the robots on the wire to which they are mapped

the robot workspace onto the wires on which the robots are constrained to move. A final projection step ensures that the locational cost subject to the motion constraints is minimized. The motion that results by the application of the derived algorithm minimizes the constrained locational cost, thus solving the constrained coverage control problem. The derived control algorithm is tested in simulation and deployed on a team of mobile robots.